\documentclass{article}
\usepackage{ijcai22}

\usepackage{times}
\usepackage{soul}
\usepackage{url}
\usepackage[hidelinks]{hyperref}
\usepackage[utf8]{inputenc}
\usepackage[small]{caption}
\usepackage{graphicx}
\usepackage{amsmath}
\usepackage{amsthm}
\usepackage{booktabs}
\usepackage[utf8]{inputenc} % allow utf-8 input
\usepackage[T1]{fontenc}    % use 8-bit T1 fonts
\usepackage{hyperref}       % hyperlinks
\usepackage{url}            % simple URL typesetting
\usepackage{booktabs}       % professional-quality tables
\usepackage{amsfonts}       % blackboard math symbols
\usepackage{nicefrac}       % compact symbols for 1/2, etc.
\usepackage{microtype}      % microtypography
\usepackage{xcolor}
\usepackage{graphicx}
\usepackage{caption}
\usepackage{subcaption}
\usepackage[ruled,linesnumbered]{algorithm2e}
\usepackage{tikz}
\usepackage[export]{adjustbox}
\usetikzlibrary{positioning,angles,quotes}
\usepackage{mathtools}
\usepackage{appendix}
\usepackage{stfloats}

\newtheorem{definition}{Definition}[section]

\newtheorem{proposition}{Proposition}[section]

\newcommand{\pro}{\text{pro}}
\newcommand{\adv}{\text{adv}}
\newcommand{\ora}{\text{ora}}

\title{Robust Reinforcement Learning as a Stackelberg Game via \\Adaptively-Regularized Adversarial Training}

\author{
Peide Huang\footnote{Contact Author}\and
Mengdi Xu\and
Fei Fang\And
Ding Zhao\\
\affiliations
Carnegie Mellon University\\
\emails
\{peideh, mengdixu\}@andrew.cmu.edu,
feif@cs.cmu.edu,
dingzhao@cmu.edu
}

\begin{document}

\maketitle

\begin{abstract}
Robust Reinforcement Learning (RL) focuses on improving performances under model errors or adversarial attacks, which facilitates the real-life deployment of RL agents. Robust Adversarial Reinforcement Learning (RARL) is one of the most popular frameworks for robust RL. However, most of the existing literature models RARL as a zero-sum simultaneous game with Nash equilibrium as the solution concept, which could overlook the sequential nature of RL deployments, produce overly conservative agents, and induce training instability. In this paper, we introduce a novel sequential formulation of robust RL \textemdash~a general-sum Stackelberg game model called RRL-Stack \textemdash~to formalize the sequential nature and provide extra flexibility for robust training. We develop the Stackelberg Policy Gradient algorithm to solve RRL-Stack, leveraging the Stackelberg learning dynamics by considering the adversary's response. Our method generates challenging yet solvable adversarial environments which benefit RL agents' robust learning. Our algorithm demonstrates better training stability and robustness against different testing conditions in the single-agent robotics control and multi-agent highway merging tasks.
\end{abstract}

\section{Introduction}

Deep reinforcement learning (DRL) has demonstrated great potential in handling complex tasks. However, its real-life deployments are hampered by the commonly existing discrepancies between training and testing environments, e.g., uncertain physical parameters in robotics manipulation tasks \cite{zhao2020sim,xu2020task} and changing hidden strategies of surrounding vehicles in autonomous driving scenarios \cite{ding2021multimodal,xu2021accelerated}. 
To remedy the fragility against model mismatches, recent advances in robust reinforcement learning (RRL) \cite{morimoto2005robust} propose to learn robust policies that maximize the worst-case performances over various uncertainties.

One popular RRL framework is the Robust Adversarial Reinforcement Learning (RARL) \cite{pinto2017robust}, which treats environment mismatches as adversarial perturbations against the agent. RARL formulates a two-player zero-sum simultaneous game between the \textit{protagonist} who aims to find a robust strategy across environments and the \textit{adversary} who exerts perturbations. Computational methods have been proposed to solve this game and find a robust strategy for the protagonist. Despite promising empirical performances in many tasks, existing approaches under the RARL framework have three potential limitations: 1) overlook the sequential nature in the deployments of RL agents, 2) produce overly conservative agents, and 3) induce training instability. These limitations will be discussed in detail in Section \ref{sec:rarl_lim}.

\begin{figure}
\includegraphics[clip, width=\linewidth, right]{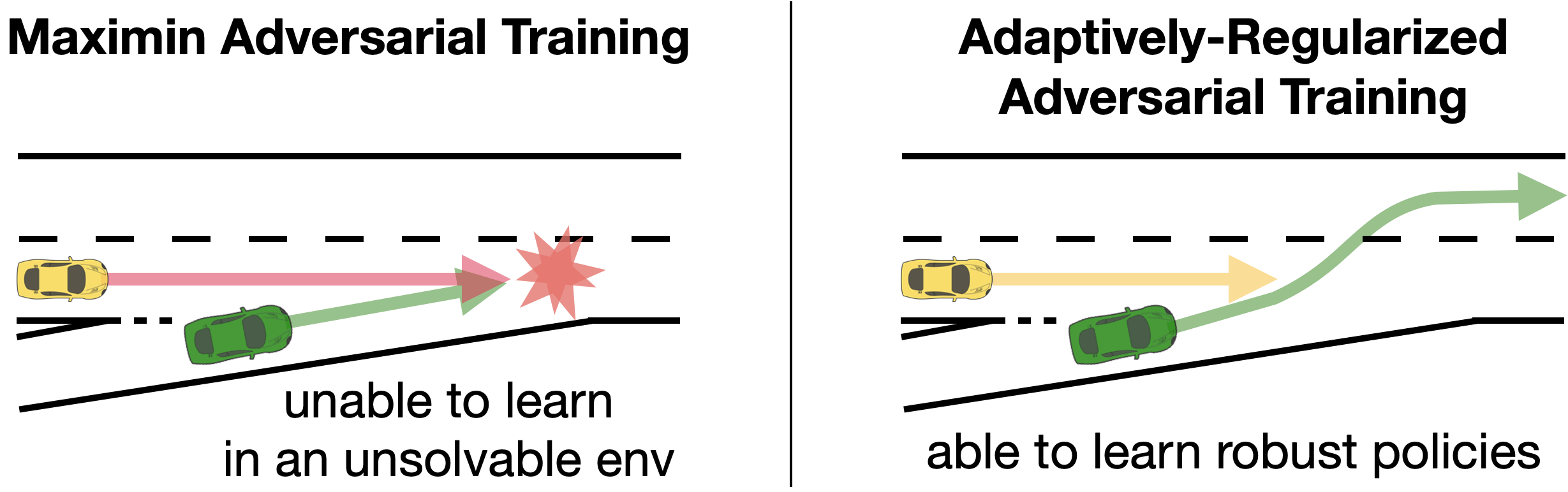}
\caption{A high-level comparison between the existing RARL formulation and our RRL-Stack formulation for robust RL. In RARL, the RL agent (green car) is trained with an adversary (yellow car) that generates extremely challenging and even unsolvable environments. In RRL-Stack, the RL agent is trained with an adaptively-regularized adversary that generates challenging yet solvable environments to improve robustness against different testing environments.}
\vspace{-0.45cm}
\end{figure}

In this paper, we propose a novel robust RL formulation -- \textit{Robust Reinforcement Learning as a Stackelberg Game (RRL-Stack)} to address these limitations. To formalize the sequential structure that the protagonist is trained first and then deployed in uncertain environments, we model robust RL as a Stackelberg game where the protagonist is the leader, and the adversary is the follower. 
By assuming the adversary optimizes a linearly combined objective of two extreme scenarios for the protagonist, RRL-Stack enables the protagonist to learn a robust policy in challenging yet solvable adversarial environments. 
RRL-Stack further provides extra flexibility to control the protagonist's conservativeness or accommodate more general multi-objective training settings.
We then leverage the Stackelberg learning \cite{fiez2020implicit} and develop the Stackelberg Policy Gradient (Stack-PG) method which has known convergence and stability guarantees to find the local surrogate solutions of the RRL-Stack.

In summary, our main contributions are: a) introducing RRL-Stack, a novel Stackelberg game-theoretical formulation of RRL (Section \ref{sec: RRL-Stack}), b) developing a Stack-PG, a policy gradient method for solving the game (Section \ref{sec: Stack-PG}), c) demonstrating that RRL-Stack formulation together with Stack-PG reduces the training instability significantly compared with existing methods, and agents learn robust but not overly conservative policies from challenging yet solvable adversarial environments (Section \ref{sec: experiments}). 

\section{Preliminaries}\label{sec: preliminaries}
\paragraph{Markov Decision Process.} We consider a Markov Decision Process (MDP) defined by the 5-tuple $(\mathcal{S}, \mathcal{A}, \mathcal{P}, r, \gamma, \mu_0)$, where $\mathcal{S}$ is a set of states, $\mathcal{A}$ is a set of continuous or discrete actions. $\mathcal{P}: \mathcal{S} \times \mathcal{A}  \rightarrow \Delta (\mathcal{S})$ is the transition probability ($\Delta (\mathcal{S})$ is the distribution over $\mathcal{S}$), $r: \mathcal{S} \times \mathcal{A} \rightarrow \mathbb{R}$ is the reward function, $\gamma$ is the discount factor and $\mu_0$ is the initial state distribution. The goal of RL is to find a policy $\pi_\theta: \mathcal{S} \times \mathcal{A} \rightarrow \mathbb{R}$ parametrized by $\theta$ that maximizes the expected return $\mathbb{E}_{\tau \sim \pi_\theta}\left[R\left(\tau\right)\right] \vcentcolon= \mathbb{E}_{\tau \sim \pi_\theta}\left[ \sum_{t=0}^{T-1} \gamma^{t} r\left(s_{t}, a_{t}\right)\right]$, where $\tau$ denotes the trajectories sampled using the policy $\pi_\theta$. For simplicity, we overload the symbol for policies as the policy parameters $\mathbb{E}_{\tau \sim \theta}\left[R\left(\tau\right)\right] \vcentcolon= \mathbb{E}_{\tau \sim \pi_\theta}\left[R\left(\tau\right)\right]$.

\paragraph{Stochastic Game.} A two-player stochastic game \cite{shapley1953stochastic} is defined by a tuple $\left(\mathcal{S}, \mathcal{A}_{1}, \mathcal{A}_{2}, \mathcal{P}, r_1, r_2, \gamma, \mu_{0}\right)$, where $\mathcal{S}$ is a set of states, $\mathcal{A}_{1}$ and $\mathcal{A}_{2}$ are action spaces of Agent 1 and 2, respectively. $\mathcal{P}: \mathcal{S} \times \mathcal{A}_1 \times \mathcal{A}_2  \rightarrow \Delta (\mathcal{S})$ is the transition probability. $r_{1}, r_{2}: \mathcal{S} \times \mathcal{A}_1 \times \mathcal{A}_2 \rightarrow \mathbb{R}$ is the reward function for Agent 1 and Agent 2 respectively. $\mu_0$ is the initial state distribution. If $r_{1}=-r_{2}$, the stochastic game is zero-sum, otherwise general-sum.

\section{Related Work}
\subsection{Robust Reinforcement Learning via Adversarial Training}\label{sec:rarl}

The transition model in training and testing could sometimes be different in RL applications. For example, in Sim2Real transfer, the physical simulator has unavoidable modeling errors, and the real world always has unexpected uncertainty \cite{akkaya2019solving}.
To address this issue, \cite{pinto2017robust} proposes Robust Adversarial Reinforcement Learning (RARL), which introduces an adversary to apply perturbations to the environment dynamics, such as disturbing forces applied to the robot's joints, gravity constants, and friction coefficients. 

RARL formulates a simultaneous zero-sum stochastic game. More concretely, let $\theta$ be the policy parameters of the protagonist who acts in an uncertain environment and $\psi$ be the policy parameters of the adversary who controls the environment dynamics. RARL assumes that the protagonist maximizes $\mathbb{E}_{\tau \sim \theta, \psi}\left[R\left(\tau\right)\right]$ while the adversary minimizes it.
\cite{tessler2019action} introduces Noisy Robust MDP to apply perturbations to the commanded action without access to the environment simulator. \cite{kamalaruban2020robust} adapts Langevin Learning Dynamics to approach robust RL from a sampling perspective. In addition, most existing works use gradient-descent-ascent-based algorithms to train agents.

\subsection{Limitations of Existing Methods}\label{sec:rarl_lim}
Despite the empirical success in many tasks, there are a few limitations in the existing formulation.
\paragraph{Overlooking the sequential nature.} Most of the existing methods use Nash equilibrium (NE) as the solution concept, which means no agent can achieve higher rewards by deviating from the equilibrium strategy. While NE is a standard solution concept for simultaneous-move games,
using NE for RRL overlooks the sequential order of actual policy deployments, %actual application and deployment of robust agents, 
as the protagonist's policy is chosen first and then tested in a range of uncertain environments \cite{jin2020local}.
Considering this sequential nature, a more appropriate solution concept is Stackelberg Equilibrium (SE). 
Although in the many zero-sum games, NE also coincides with SE, it is not the case in deep RARL as discussed by \cite{tessler2019action,zhang2019multi}.
One of the reasons is that with policy parametrizations in Deep RL, e.g., using neural networks, the objectives of stochastic games are nonconvex-nonconcave in the parameter space, and as a result, maximin value is not equal to minimax value in general.
Different from most existing works, our RRL-Stack adopts a sequential
game-theoretical formulation and solution concept, which formalizes the sequential order of actual policy deployments.

\paragraph{Producing overly conservative agents.} Existing work often assumes that the protagonist and adversary play a zero-sum game where the adversary minimizes the protagonist's expected return. However, such a formulation encourages the adversary to generate extremely difficult, even unsolvable environments for the protagonist. As a result, the protagonist may choose an overly conservative strategy
or even not be able to learn any meaningful policies because the most adversarial environment is completely unsolvable. 
\cite{dennis2020emergent} proposes PAIRED to mitigate this issue by replacing the return in the objective with the minimax regret, which has the closest relation to our proposed method. The protagonist's regret is defined as the difference between the maximum possible return and the current return, which can be seen as a special case of our adversary's objective. Some other literature \cite{shen2020deep} dealing with state-robustness uses $l_p$ ball to constrain the adversarial perturbation on the state.

\paragraph{Potential training instability.} Besides the two aforementioned issues related to the game formulation, the gradient-descent-ascent learning dynamics in RARL could lead to significant training instability even in the simple linear-quadratic system. The unstable training is partially due to the non-stationarity of environments, as the environment controlled by the adversary could be constantly evolving.
Recent research in RARL \cite{zhang2020stability,yu2021robust} aims to tackle this problem, yet only for linear-quadratic systems with the zero-sum formulation. In contrast, our algorithm relying on the Stackelberg learning dynamics \cite{fiez2020implicit} can be applied to more general RL settings.

\section{Method}
We first introduce our proposed formulation, Robust Reinforcement Learning as a Stackelberg Game (RRL-Stack) in Section \ref{sec: RRL-Stack}. We then present our main algorithm, Stackelberg Policy Gradient (Stack-PG) in Section \ref{sec: Stack-PG}.

\subsection{Robust RL as a Stackelberg Game}\label{sec: RRL-Stack}
Although formulated as simultaneous games in most of the existing works, adversarial and robust training are in fact sequential \cite{jin2020local}. In robust RL via adversarial training, 
the protagonist has to choose its policy first and then the adversary chooses the best responding policy given the protagonist's policy. To formalize this inherently sequential structure, we formulate robust RL as a Stackelberg game as follows:
\vspace{-0.1cm}
\begin{align*}
\max_{\theta \in \Theta} \mathbb{E}_{\tau \sim \theta, \psi}\left[R_\pro\left(\tau\right)\right] \quad s.t. \\ 
\psi \in \arg \max _{\psi^{\prime} \in \Psi} \mathbb{E}_{\tau \sim \theta, \psi^{\prime}}\left[-R_\pro\left(\tau\right)\right],
\end{align*}
where $\theta$ and $\psi$ parametrize the protagonist's and the adversary's policy, respectively. $\mathbb{E}_{\tau \sim \theta, \psi}\left[R_\pro\left(\tau\right)\right]$ is the expected return of the protagonist.

With the sequential game structure, we now aim to address the limitations caused by the widely adapted zero-sum formulation. 
It could result in an overly conservative protagonist's policy or even disable the protagonist's effective learning since the adversary is encouraged to produce a difficult, even completely unsolvable environment.
To address this limitation, 
we introduce an oracle term that encodes the highest possible return of the protagonist in the current adversarial environment. Inspired by the alpha-maxmin expected utility in the economics literature \cite{li2019equilibrium}, we linearly combined this oracle term with the original adversary's objective. Formally, \looseness=-1

\vspace{-0.3cm}
\begin{align}\label{eq:rrl-stack}
 \max _{\theta \in \Theta} \mathbb{E}_{\tau \sim \theta, \psi}\left[R_\pro\left(\tau\right)\right]& \quad s.t. \\ 
 \psi \in \arg \max _{\psi^{\prime} \in \Psi} \alpha\mathbb{E}_{\tau \sim \theta, \psi^{\prime}}\left[-R_\pro\left(\tau\right)\right] &+ (1-\alpha) V^*(\psi^{\prime}),
\end{align}

where $V^*(\psi^{\prime})$ is the highest possible return of the protagonist given the current adversarial environment $\psi^{\prime}$. This term is approximated by training an oracle RL agent that is locally optimized. Let the oracle agent's policy be parametrized by $\omega$, $V^*(\psi^{\prime}) \vcentcolon= \sup_{\omega} \mathbb{E}_{\tau \sim \omega, \psi^{\prime}}\left[R_\ora \left(\tau\right)\right]$.

This term can be seen as a regularization of the adversary's objective. The adversary is incentivized to generate environments that are challenging to solve for the current protagonist but solvable for the oracle agent. As the protagonist learns to solve the current environment, the adversary is forced to find harder environments to receive a higher reward, adaptively increasing the difficulty of the generated environments.

The coefficient $\alpha \in [0, 1]$ balances how adversarial the environment is and linearly combines two extreme scenarios:
\begin{itemize}
  \item When $\alpha=1$, RRL-Stack's solution corresponds to the \textit{Maximin} robust strategy.
  \item When $\alpha=0.5$, the solution corresponds to the strategy when the adversary maximizes the protagonist's regret, which is defined by the difference between the maximum possible return and the current return.
  %while the protagonist is maximizing its own cumulative reward.
  \item When $\alpha=0$, the solution corresponds to the \textit{Maximax} strategy, where the protagonist chooses a strategy that yields the best outcome in the most optimistic setting.
\end{itemize}

Denote $f_{\pro}\left(\theta, \psi\right) \vcentcolon= \mathbb{E}_{\tau \sim \theta, \psi}\left[R_\pro\left(\tau\right)\right]$ and $f_{\adv}\left(\theta, \psi\right) \vcentcolon= \alpha\mathbb{E}_{\tau \sim \theta, \psi^{\prime}}\left[-R_\pro\left(\tau\right)\right] + (1-\alpha) V^*(\psi^{\prime})$. The solution to the Stackelberg game is the Stackelberg Equilibirium. We have the following definition:
\begin{definition} [Stackelberg Equilibrium]\textbf{(SE)}
  The joint strategy $(\theta^*, \psi^*) \in \Theta \times \Psi$ of the protagonist is a Stackelberg equilibrium if
  \begin{equation*}
  \inf _{\psi \in \mathcal{R}\left(\theta^{*}\right)} f_{\pro}\left(\theta^{*}, \psi\right) \geq \inf _{\psi \in \mathcal{R}\left(\theta\right)} f_{\pro}\left(\theta, \psi\right), \quad \forall \theta \in \Theta,
  \end{equation*}
  where $\mathcal{R}\left(\theta\right)=\left\{\psi^{\prime} \in \Psi \mid f_{\adv}\left(\theta, \psi^{\prime}\right) \geq f_{\adv}\left(\theta, \psi\right), \forall \psi \in \Psi\right\}$ is the best response set of the adversary, and $\psi^* \in \mathcal{R}\left(\theta^*\right)$
\end{definition}

We adapt Proposition 4.4 of \cite{bacsar1998dynamic} to formalize the relationship between the return at Stackelberg Equilibrium and Nash equilibrium:

\begin{proposition}
Consider an arbitrary sufficiently smooth two-player general-sum game $(f_{\pro}, f_{\adv})$ on continuous strategy spaces.  Let $f_{\pro}^{\mathcal N *}$ denote the supremum of all Nash equilibrium return for the protagonist and $f_{\pro}^{\mathcal S}$ denote an arbitrary Stackelberg equilibrium return for the protagonist. Then, if $\mathcal{R}(\theta)$ is a singleton for every $\theta \in \Theta$, $f_{\pro}^{\mathcal S} \geq f_{\pro}^{\mathcal N *}$.
\end{proposition}

However, since one cannot expect to find the global solution of the Stackelberg game efficiently with a general non-convex-non-concave objective, we define the following local equilibrium concept using the sufficient conditions of SE.

\begin{definition} [Differential Stackelberg Equilibrium] \textbf{(DSE)} \cite{fiez2020implicit}
  The joint strategy $(\theta^*, \psi^*) \in \Theta \times \Psi$ with $\psi^* = r(\theta^*)$, where $r$ is an implicit mapping defined by $\nabla_{\psi}f_{\adv}(\theta^*, \psi^*)=0$, is a differential Stackelberg equilibrium if $D f_{\pro}(\theta^*, r(\theta^*))=0$ and $\nabla^2 f_{\pro}(\theta^*, r(\theta^*))$ is negative definite ($D(\cdot)$ denotes the total derivative).
\end{definition}

To this end, the key question is how to solve RRL-Stack. We now explain how to develop Stackelberg Policy Gradient (Stack-PG) leveraging the Stackelberg learning dynamics, which have known convergence and stability guarantees to find the DSE under sufficient regularity conditions.

\subsection{Stackelberg Policy Gradient (Stack-PG)}\label{sec: Stack-PG}

\paragraph{Stackelberg learning dynamics.} \cite{fiez2020implicit} assume that there exists an implicit mapping from $\theta$ to the best-response $\psi$. The protagonist updates its parameters based on the total derivative ($Df_{\pro} \vcentcolon= \mathrm{d} f_{\pro}\left(\theta, r^{*}\left(\theta\right)\right)/\mathrm{d} \theta$) instead of the partial derivative ($\nabla_{\theta}f_{\pro}$). Since the follower chooses the best response $\psi = r^*(\theta)$, the follower's policy is an implicit function of the leader's. The leader therefore can take the total derivative of its objective to update its policy:
\begin{equation}\label{eq:td}
\frac{\mathrm{d} f_{\pro}\left(\theta, r^{*}\left(\theta\right)\right)}{\mathrm{d} \theta}=\frac{\partial f_{\pro}\left(\theta, \psi\right)}{\partial \theta}+\frac{\mathrm{d} r^{*}\left(\theta\right)}{\mathrm{d} \theta} \frac{\partial f_{\pro}\left(\theta, \psi\right)}{\partial \psi}
\end{equation}
The implicit differentiation term can be computed using the implicit function theorem \cite{abraham2012manifolds}:
\begin{equation}\label{eq:ift}
\frac{\mathrm{d} r^{*}\left(\theta\right)}{\mathrm{d} \theta}=\left(\frac{\partial^{2} f_{\adv}\left(\theta, \psi\right)}{\partial \theta \partial \psi}\right)\left(-\frac{\partial^{2} f_{\adv}\left(\theta, \psi\right)}{\partial \psi^{2}}\right)^{-1}
\end{equation}
By combining Eq.\ref{eq:td} and Eq.\ref{eq:ift}, we obtain the updating rule for the protagonist. The Stackelberg learning dynamics provide local convergence guarantees to DSE under regularity conditions. We include a numerical example in the Appendix B \footnote{https://arxiv.org/abs/2202.09514} to demonstrate: 1) why SE is a more appropriate solution concept for robust learning than NE, and 2) how the Stackelberg learning dynamics converge to DSE while the gradient-descent-ascent algorithm fails to converge.

The computation of Hessian required by the Stackelberg learning dynamics takes complexity $O(n^2)$, where $n$ is the number of policy parameters. It could be prohibitively slow when $n$ is large. There are some techniques for efficiently computing unbiased Hessian approximations of deep neural networks such as Curvature Propagation \cite{martens2012estimating}.
The computation of the inverse of Hessian is another burden but can be alleviated by approximation methods such as conjugate gradient \cite{shewchuk1994introduction} and minimal residual \cite{saad1986gmres}.
We leave integrating efficient computation methods to future work.

\begin{algorithm}[tb]
\SetAlgoLined
\textbf{Input:} $\{\tau_{\pro}\}^{M}, \{\tau_{\ora}\}^{M}, \theta_{k-1}$, learning rate $\gamma_{\theta}$ \\
$\omega_{\mathcal{S}, \theta} \leftarrow \hat\nabla_{\theta} f_{\pro}\left(\theta, \psi\right) + \hat \nabla_{\theta}\hat\nabla_{\psi} f_{\adv}\left(\theta, \psi\right)\left(-\hat\nabla_{\psi}^{2} f_{\adv}\left(\theta, \psi\right) + \lambda I \right)^{-1} \hat\nabla_{\psi} f_{\pro}\left(\theta, \psi\right)$ \\
$\theta_{k} \leftarrow \theta_{k-1} + \gamma_{\theta} \omega_{\mathcal{S}, \theta}$ \\
% $\lambda \leftarrow \eta\lambda$ \\
\textbf{Output:} $\theta_k$ \\
\caption{StackelbergPolicyGradient (Stack-PG)}\label{alg:StackPG}
\end{algorithm}

\begin{algorithm}[tb]
\SetAlgoLined
\textbf{Input:} $\{\tau_{\pro}\}^{M}, \{\tau_{\ora}\}^{M}, \psi_{k-1}$, learning rate $\gamma_{\psi}$, $\text{auto-tuning}\in\{True, False\}$, smoothing factor $\rho$  \\
$g_1 \leftarrow \frac{\partial}{\partial \psi}\mathbb{E}\left[-R_\pro\left(\tau\right)\right], g_2 \leftarrow \frac{\partial}{\partial \psi} \mathbb{E}_{\tau \sim \omega, \psi}\left[R_\ora \left(\tau\right)\right]$ \\
\If{auto-tuning}{Find the optimal $\alpha^*$ by solving (\ref{eq:dynamic_weight})\\
$\alpha \leftarrow \rho \alpha + (1-\rho)\alpha^*$ \tcp{Moving average}}
$\omega_{\mathcal{S}, \psi} \leftarrow \alpha g_1 + (1-\alpha)g_2$ \\
$\psi_{k} \leftarrow \psi_{k-1} + \gamma_{\psi} \omega_{\mathcal{S}, \psi}$ \\
% $\lambda \leftarrow \eta\lambda$ \\
\textbf{Output:} $\psi_k$ \\
\caption{MultiPolicyGradient}\label{alg:MultiPolicyGradient}
\end{algorithm}

\paragraph{Protagonist's updating rule.} Based on the Stackelberg learning dynamics, we develop the update rule for the protagonist, Stackelberg Policy Gradient (Stack-PG), as shown in Algorithm \ref{alg:StackPG}. 
Similar to the policy gradient algorithm, we obtain unbiased estimators for the first-order and second-order gradient information based on trajectory samples. Details for the unbiased estimators are in Appendix A.
We can also incorporate state-dependent baselines into the gradient estimators to reduce variance. 
The regularization term $\lambda I$ ensures the Hessian estimate is invertible, where $\lambda$ is a scalar and $I$ is the identity matrix.
Note that as we increase the value of $\lambda$, the protagonist's update first resembles LOLA \cite{foerster2017learning} and eventually becomes the standard policy gradient.

\paragraph{Adversary's updating rule and auto-tuning $\alpha$.} 
The adversary updates its parameters with a policy-gradient-based algorithm. Since there are two terms in the adversary's objective, it can be viewed from a multi-objective RL perspective. 
Instead of manually tuning the value of $\alpha$, we can dynamically update it using the multiple-gradient descent algorithm (MGDA) \cite{desideri2012multiple} for multi-objective learning. Let $g_1 \vcentcolon= \frac{\partial}{\partial \psi}\mathbb{E}_{\tau \sim \theta, \psi}\left[-R_{\pro}\left(\tau\right)\right], g_2 \vcentcolon= \frac{\partial}{\partial \psi} \mathbb{E}_{\tau \sim \omega, \psi}\left[R_\ora \left(\tau\right)\right]$. We want to find the $\alpha^*$ that approximately maximizes the minimal improvement of the two terms by solving the optimization problem:
\begin{equation}\label{eq:dynamic_weight}
\min_{\alpha} \frac{1}{2}\left\|\alpha g_1 + (1 - \alpha) g_2\right\|^{2}, \text { s.t. }  \alpha \in [0, 1].
\end{equation}
It is the dual form of the primal optimization problem:
\begin{equation}
\max _{\|d\| \leq 1} \min _{i}\left\langle d, g_{i}\right\rangle,
\end{equation}
where the optimal $d^{*}=(\alpha g_1 + (1 - \alpha) g_2) / \lambda$ and $\lambda$ is the Lagrangian multiplier of the constraints $\|d\| \leq 1$. After solving for $\alpha^*$, we use exponential moving average to update $\alpha$ smoothly. The updating algorithm of the adversary is shown in Algorithm \ref{alg:MultiPolicyGradient}. Note that this $\alpha$ auto-tuning is not required but is a tool for automatic hyper-parameter tuning.

\paragraph{Oracle agent's updating rule.} The oracle agent can be trained using any on- or off-policy optimization algorithm. In practice, we find that performing multiple policy optimization steps for the oracle agent in each iteration usually serves the purpose well.\looseness=-1

\begin{algorithm}[tb]
  \SetAlgoLined
   \textbf{Input:} Protagonist's policy $\pi_{\theta}$, Adversary's policy $\pi_{\psi}$, Oracle Agent's policy $\pi_{\omega}$, Number of trajectories $M$\\
   Initialize learnable parameters $\theta_0, \psi_0, \omega_0$\\
   \For{$k=1,2, \ldots, N_{\text{iter}}$}{
      % $\theta_{k} \leftarrow \theta_{k-1}$\\
      $\{\tau_{\pro}\}^{M} \leftarrow \text{rollout}(\pi_{\theta_{k}}, \pi_{\psi_{k-1}})$ \\
      $\{\tau_{\ora}\}^{M} \leftarrow \text{rollout}(\pi_{\omega}, \pi_{\psi_{k-1}})$ \\
      $\theta_{k} \leftarrow \text{StackelbergPolicyGradient}(\{\tau_{\pro}\}^{M}, \{\tau_{\ora}\}^{M})$ \\
      % $\psi_{k} \leftarrow \psi_{k-1}$\\
      % $\{\tau_{\pro}\}^{M} \leftarrow \text{rollout}(\pi_{\theta_{k}}, \pi_{\psi_{k-1}})$ \\
      % $\{\tau_{\ora}\}^{M} \leftarrow \text{rollout}(\pi_{\ora}, \pi_{\psi_{k-1}})$ \\
      $\psi_{k} \leftarrow \text{MultiPolicyGradient}(\{\tau_{\pro}\}^{M}, \{\tau_{\ora}\}^{M})$ \\
      % $\pi_\ora \leftarrow \beta \pi_\ora + (1-\beta)\pi_{\theta_{k}}$ \\ 
      Train $\pi_\omega$ in the environment given by $\psi_k$
   }
  \textbf{Output:} $\theta_{N_{\text{iter}}}$\\
  % and $\psi_{N_{\text{iter}}}$
  \caption{Solving RRL-Stack with Stack-PG}\label{alg:RRL-Stack}
\end{algorithm}

The Algorithm \ref{alg:RRL-Stack} summarizes our main algorithm. At each iteration, we first rollout trajectories using the protagonist's and adversary's policy (Line 4), as well as trajectories using the oracle agent's and adversary's policy (Line 5). Next, we use the Stack-PG to update the protagonist's policy parameters (Line 6). Then we use the policy-gradient-based method to update the adversary's policy parameters (Line 7). Finally, we train the oracle agent in the current adversarial environment till reaching local convergence (Line 8).

\section{Experiments} \label{sec: experiments}
\begin{figure*}[!htp]
  \centering
  \includegraphics[clip, width=\linewidth]{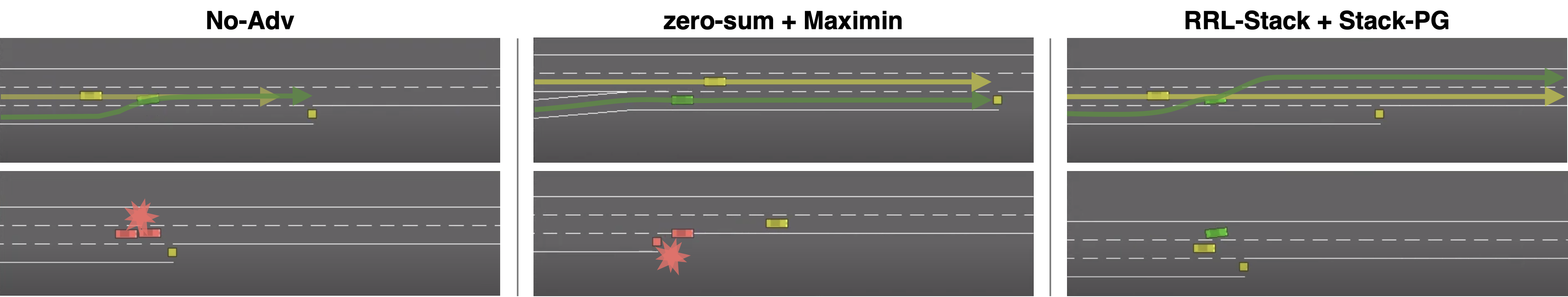}
  \vspace{-0.5cm}
  \caption{Highway merging policy visualization during testing. The green car is controlled by the protagonist and the yellow car is controlled by the adversary. Cars turning red mean collisions. The opaque lines represent the driving trajectories. The upper row contains moments close to the beginning of the episode, and the lower row contains moments close to the end of the episode. 
  }
  \label{fig:highway_vis}
%   \vspace{-0.5cm}
\end{figure*}
We conduct experiments to answer the following questions: \textbf{(Q1)} Does our method produce challenging yet solvable environments? \textbf{(Q2)} Does our method improve the robustness and training stability? \textbf{(Q3)} How does the choice of $\alpha$ influence the performance of the protagonist?

\subsection{Benchmark Algorithms} \label{sec:existing alg}
We consider 2 game formulations: zero-sum and RRL-Stack, and 3 existing learning algorithms as follows:

\paragraph{Gradient-descent-ascend (GDA).} GDA alternates between the policy gradient updates of protagonist and adversary at 1:1 ratio \cite{zhang2021don}.

\paragraph{Maximin operator.} Maximin operator is similar to GDA, but the difference is that the adversary updates multiple iterations between each update of the protagonist \cite{tessler2019action}. In the experiments, we use a ratio of 1:3 to alternate between the updates of the protagonist and the adversary.

\paragraph{Learning with opponent-learning awareness (LOLA).} LOLA \cite{foerster2017learning} is a seminal work in considering the opponent while doing gradient ascent. 
We choose LOLA for its similarity to the Stackelberg learning dynamics.

We consider several combinations of game formulations and learning algorithms: zero-sum game formulation with GDA, Maximin operator, and LOLA; RRL-Stack game formulation with GDA, Maximin operator, and Stack-PG. \textit{Zero-sum + GDA} and \textit{zero-sum + Maximin operator} are widely used in existing works of RARL. We also include a non-robust training baseline (\textit{No-Adv}) to highlight the difference between robust and non-robust training.

Without specific mention, the policies are parametrized by MLPs with two hidden layers. All the agents are trained using policy gradient algorithms with Adam optimizer and the same learning rate. Each plot is computed with 5 policies generated from different random seeds. The episodic reward is evaluated over 48 episodes for each policy. 
More experiment details are included in Appendix C.

\subsection{Highway Merging Task}

In highway merging tasks \cite{highway-env}, the protagonist aims to control the ego vehicle (green) to merge into the main lane while avoiding collision with the other yellow vehicle or hitting the end of the ramp. At every timestep, the adversary controls the aggressiveness of the yellow vehicle whose acceleration is proportional to the aggressiveness. The yellow vehicle can only drive in the middle lane, while the ego vehicle can switch lanes. 

In this experiment, we aim to answer \textbf{(Q1)} and \textbf{(Q2)}. We compare our method \textit{RRL-Stack + Stack-PG} with $\alpha=0.5$, against the benchmark method \textit{zero-sum + Maximin operator} and non-robust training (\textit{No-Adv}) agents. To evaluate the robustness against different environment parameters, we vary the aggressiveness of the yellow vehicle from 0 to 10 and compare the episodic reward of each method.

\begin{figure}[!b]
  \centering
  \vspace{-0.3cm}
  \includegraphics[width=0.8\linewidth]{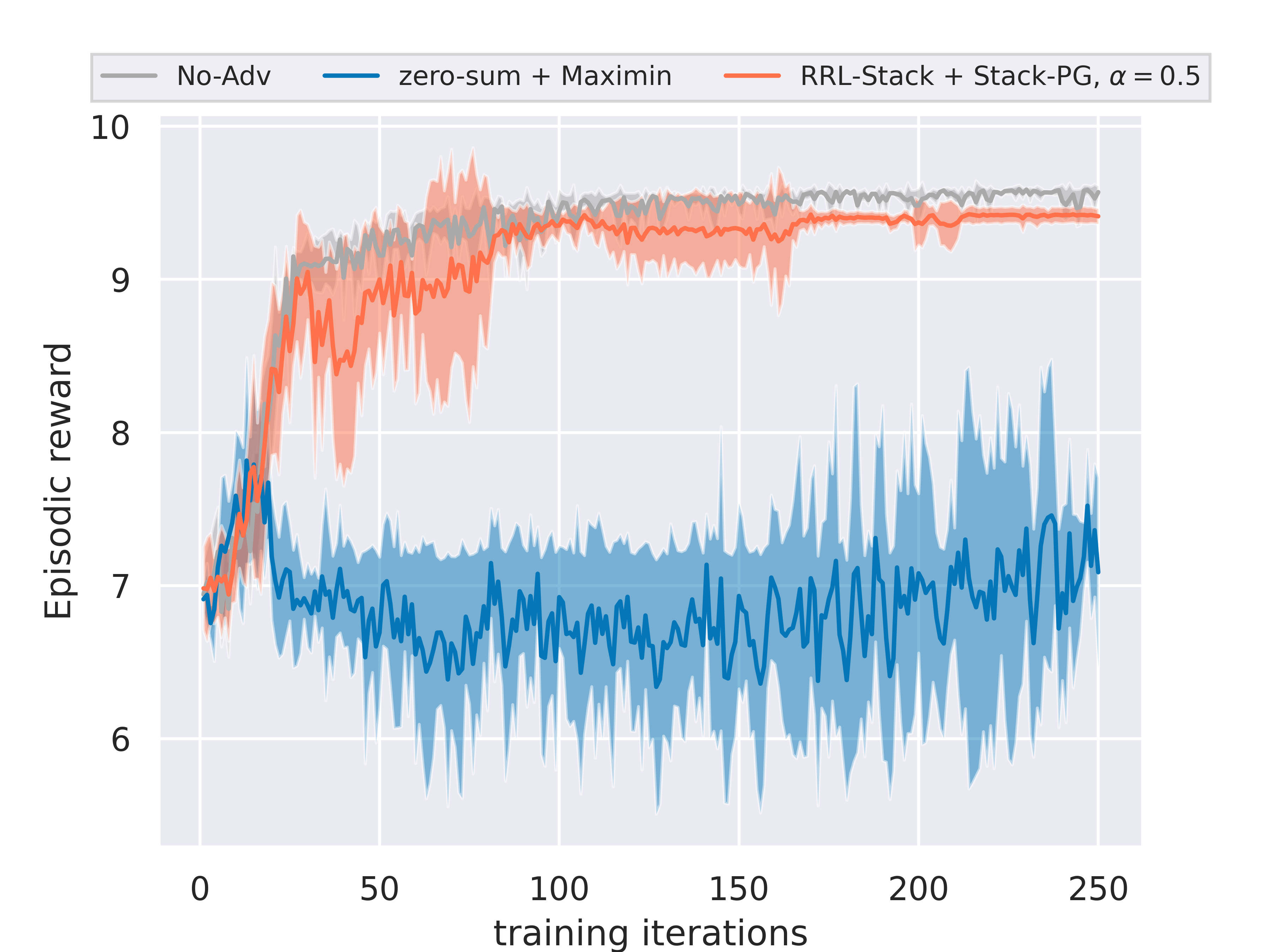}
  \vspace{-0.20cm}
  \caption{Training curve of highway merging. The x-axis is the training iterations of the protagonist. The y axis is the episodic reward evaluated in the environment without an adversary. The shaded area represents the standard deviation}\label{fig:HighwayTraining}
  \vspace{-0.30cm}
\end{figure}

To answer \textbf{(Q1)} about whether \textit{RRL-Stack + Stack-PG} produces challenging yet solvable environments and allows the protagonist to learn robust policies, we visualize the trajectories of the final policies in Fig.~\ref{fig:highway_vis}.
For \textit{No-Adv}, since the yellow car does not collide with the protagonist during the training, the protagonist is unaware of the danger of the yellow car. Therefore, \textit{No-Adv} exhibits poor robustness during testing in unseen environments. For \textit{zero-sum + Maximin operator}, the adversary quickly finds the policy to keep blocking the main lane before the ego vehicle enters the lane, which makes the environment completely unsolvable for the protagonist. In this case, the protagonist cannot learn any robust policies but only hits the end of the ramp.

In contrast, for \textit{RRL-Stack + Stack-PG} with $\alpha=0.5$, since the adversary is not fully adversarial but maximizing the regret of the protagonist, the resulting environments are challenging but still solvable. The protagonist learns robust policies to switch to the middle lane and immediately switch to the leftmost lane to avoid potential collisions. Therefore, \textit{RRL-Stack + Stack-PG} agents exhibit more robustness to the unseen environments than the baselines.

To answer \textbf{(Q2)} about whether our method improves the robustness and training stability, Fig.~\ref{fig:HighwayTraining} shows the rewards through the training process. The rewards are evaluated in the same environment without the adversary for a fair comparison. The non-robust training (\textit{No-Adv}) is stable and converges fast because the protagonist is training and evaluating both in the environment without the adversary (however we will observe it is not robust against unseen environments). We observe that after around 20 iterations, the adversary of \textit{zero-sum + Maximin operator} quickly learns to make the task unsolvable, so the protagonist's reward is driven to the lowest possible value for the rest of the training. In contrast, \textit{RRL-Stack + Stack-PG} adjusts the difficulty of the environment adaptively to ensure the task remains solvable and the protagonist keeps learning robust policies. 

\begin{figure}[t]
  \centering
%   \vspace{-0.45cm}
  \includegraphics[trim=0 0 0 12, clip, width=0.8\linewidth]{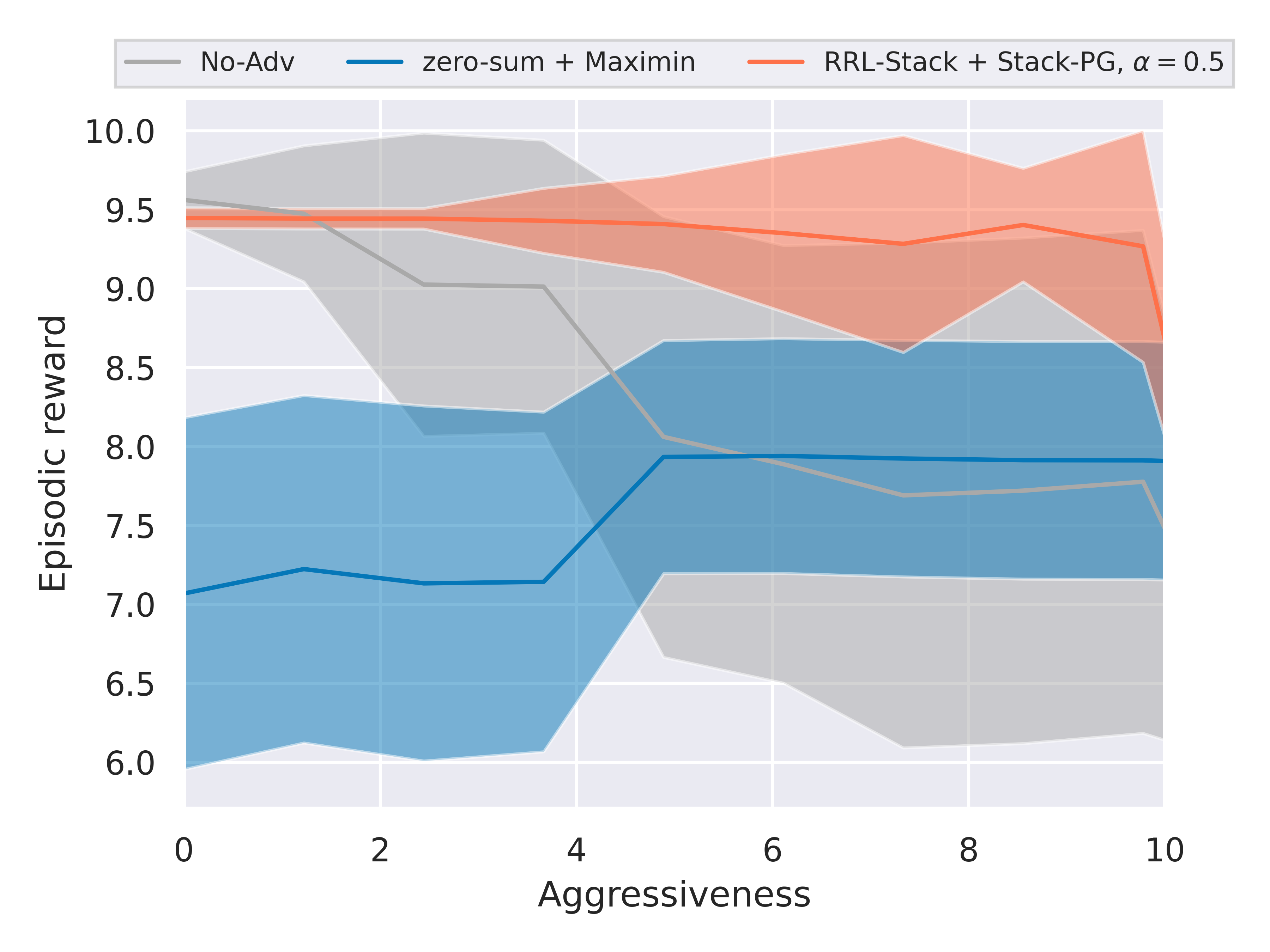}
  \vspace{-0.45cm}
  \caption{Robustness against different aggressiveness levels. The shaded area represents the standard deviation. }\label{fig:HighwayRobustness}
  \vspace{-0.3cm}
\end{figure}
% \vspace{-0.1cm}
Fig.~\ref{fig:HighwayRobustness} shows the rewards against different aggressiveness levels. We observe that policies trained with \textit{No-Adv} and \textit{zero-sum + Maximin operator} fail to be robust against different aggressiveness levels. In contrast, \textit{RRL-Stack + Stack-PG} agents are much more robust against unseen environment parameters during testing. The mean reward of \textit{RRL-Stack + Stack-PG} outperforms the baselines by a large margin, and the variance is significantly smaller.

\subsection{LunarLander with Actuation Delay}

\begin{figure}[b]
  \centering
  % left bottom right top
  \includegraphics[trim=22 0 45 0, clip, width=1.0\linewidth]{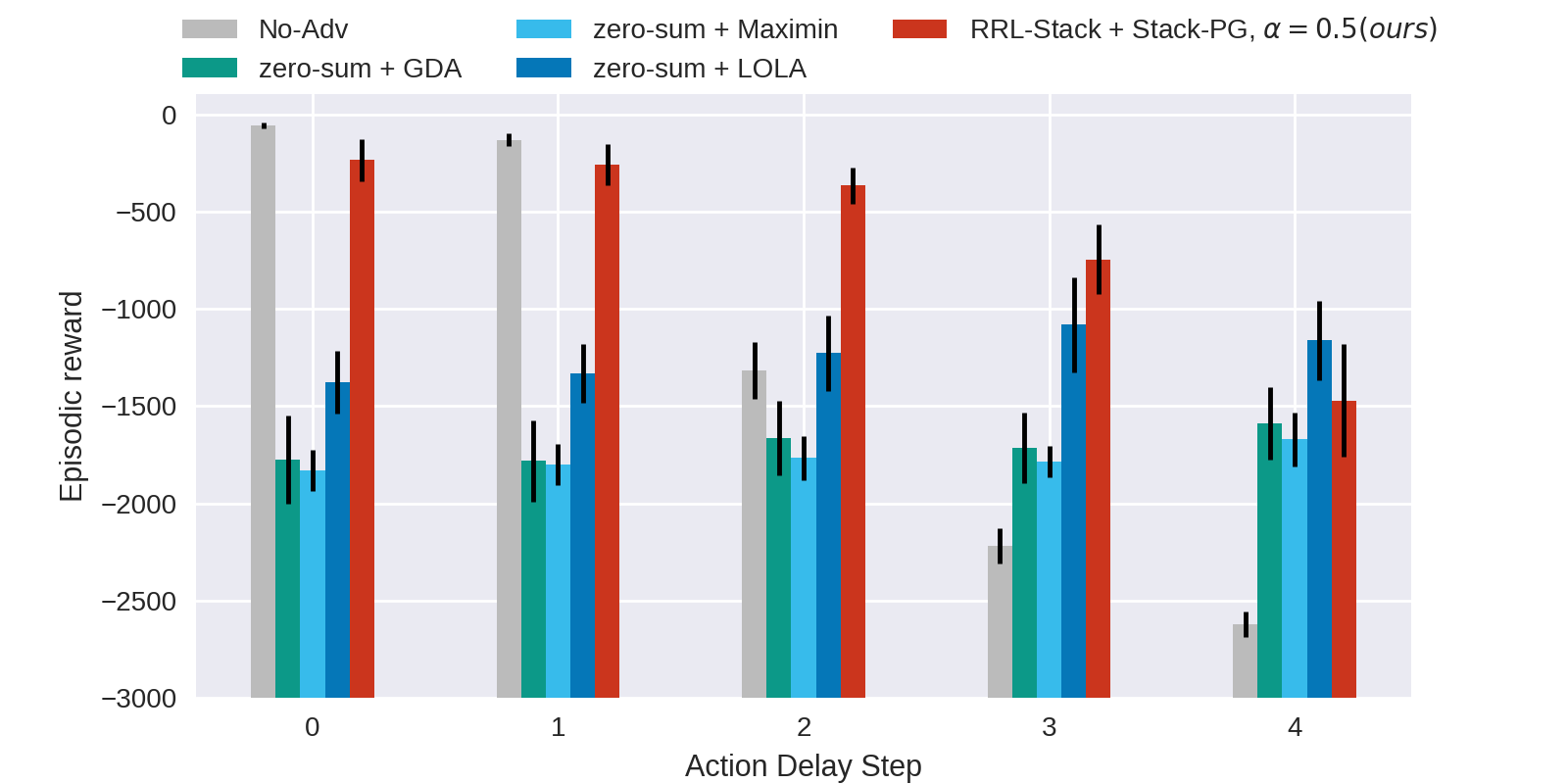}
  \vspace{-0.65cm}
  \caption{Episodic return with different action delay steps. The black error bar indicates the standard deviation. }\label{fig:LunarLanderRobustness}
  \vspace{-0.4cm}
\end{figure}

Actuation delay is a common problem in robotic control \cite{chen2021delay}. We modify the LunarLander environment in OpenAI Gym \cite{brockman2016openai} to simulate the effects of actuation delay. The protagonist in LunarLander has 4 discrete actions: shut off all engines, turn on the left engine, turn on the right engine and turn on the main engine. The objective is to train a protagonist that is robust to the commanded action being delayed to execute by several steps. During the adversarial training, at each time step, the adversary chooses a number from $\{0, 1, 2, ..., 10\}$, representing the delay steps of the protagonist's action. During testing, the delay step is fixed throughout each episode. 
% \vspace{-0.20cm}

% \vspace{-0.20cm}
In Fig.~\ref{fig:LunarLanderRobustness}, we show the episodic reward against different action delay steps, compared with the baselines. Only action delay steps from $0$ to $4$ are shown here since the returns of more delay steps are low for all methods and not meaningful statistically. We find that \textit{RRL-Stack + Stack-PG} outperforms the baselines, particularly at delay step $0$ to $3$. 

\begin{figure}[t]
  \centering
%   \vspace{-0.1cm}
  \includegraphics[trim=22 0 45 20, clip, width=1.0\linewidth]{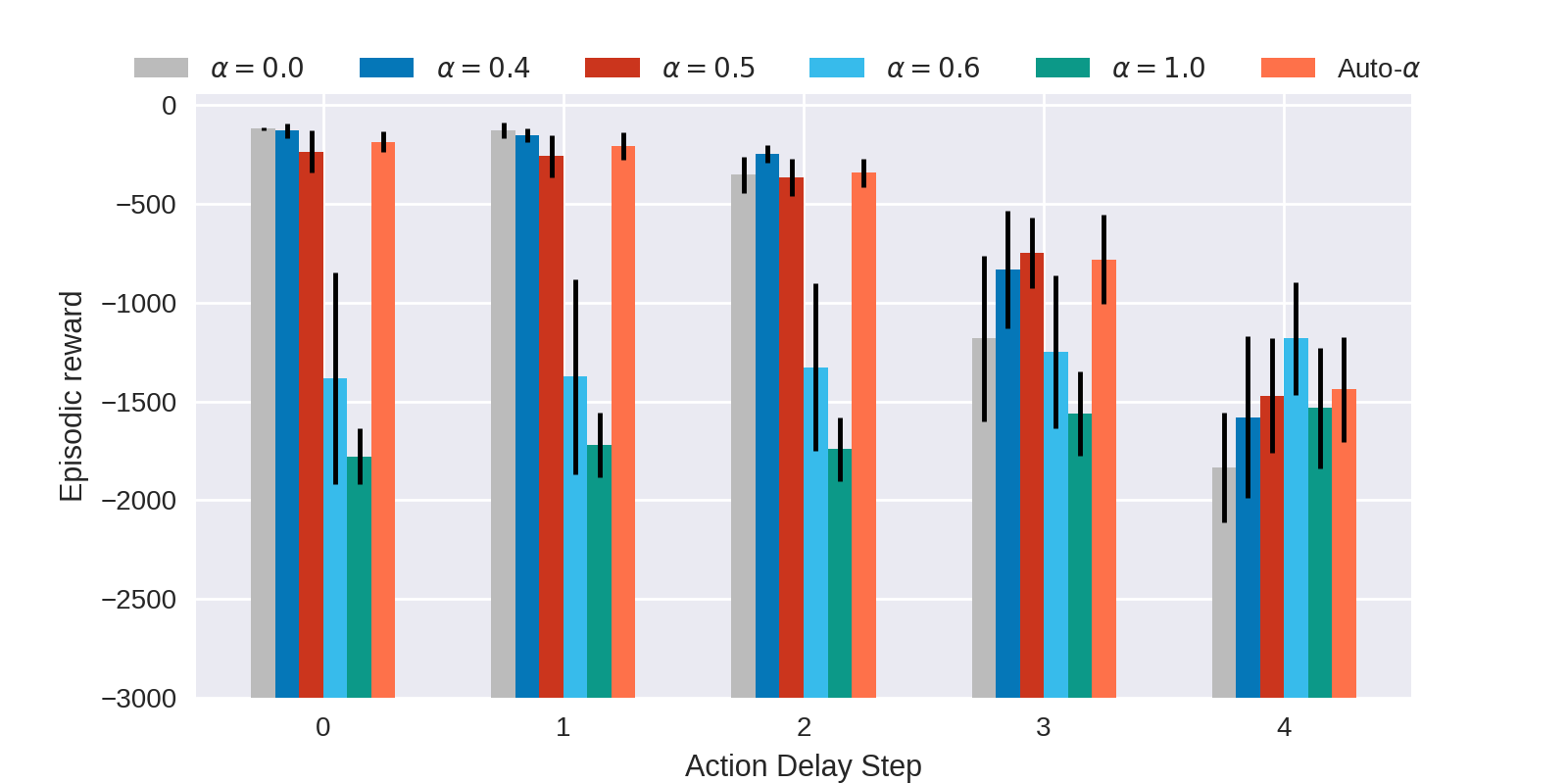}
  \vspace{-0.65cm}
  \caption{Effects of different $\alpha$ on the episodic rewards.}\label{fig:LunarLanderRobustnessAlpha}
  \vspace{-0.50cm}
\end{figure}

To answer \textbf{(Q3)} about the effects of $\alpha$ on the robustness of \textit{RRL-Stack + Stack-PG}, we study the robustness of $\alpha \in \{0.0, 0.4, 0.5, 0.6, 1.0\}$ as well as our auto-tuning $\alpha$ (\textit{Stack+Auto-$\alpha$}) in Fig.~\ref{fig:LunarLanderRobustnessAlpha}. When $\alpha=0.4, 0.5$, the protagonist maintains highly robust against different action delay steps, while $\alpha=0.0$ results in non-robust agents and $\alpha=0.6, \alpha=1.0$ produce overly-conservative agents. With \textit{Stack+Auto-$\alpha$}, the protagonist achieves comparable performance to the best performing $\alpha=0.4, 0.5$ without fine-tuning.

% \vspace{-0.50cm}

\begin{figure}[h]
  \vspace{-0.15cm}
  \centering
  \includegraphics[trim=22 0 45 5, clip, width=1.0\linewidth]{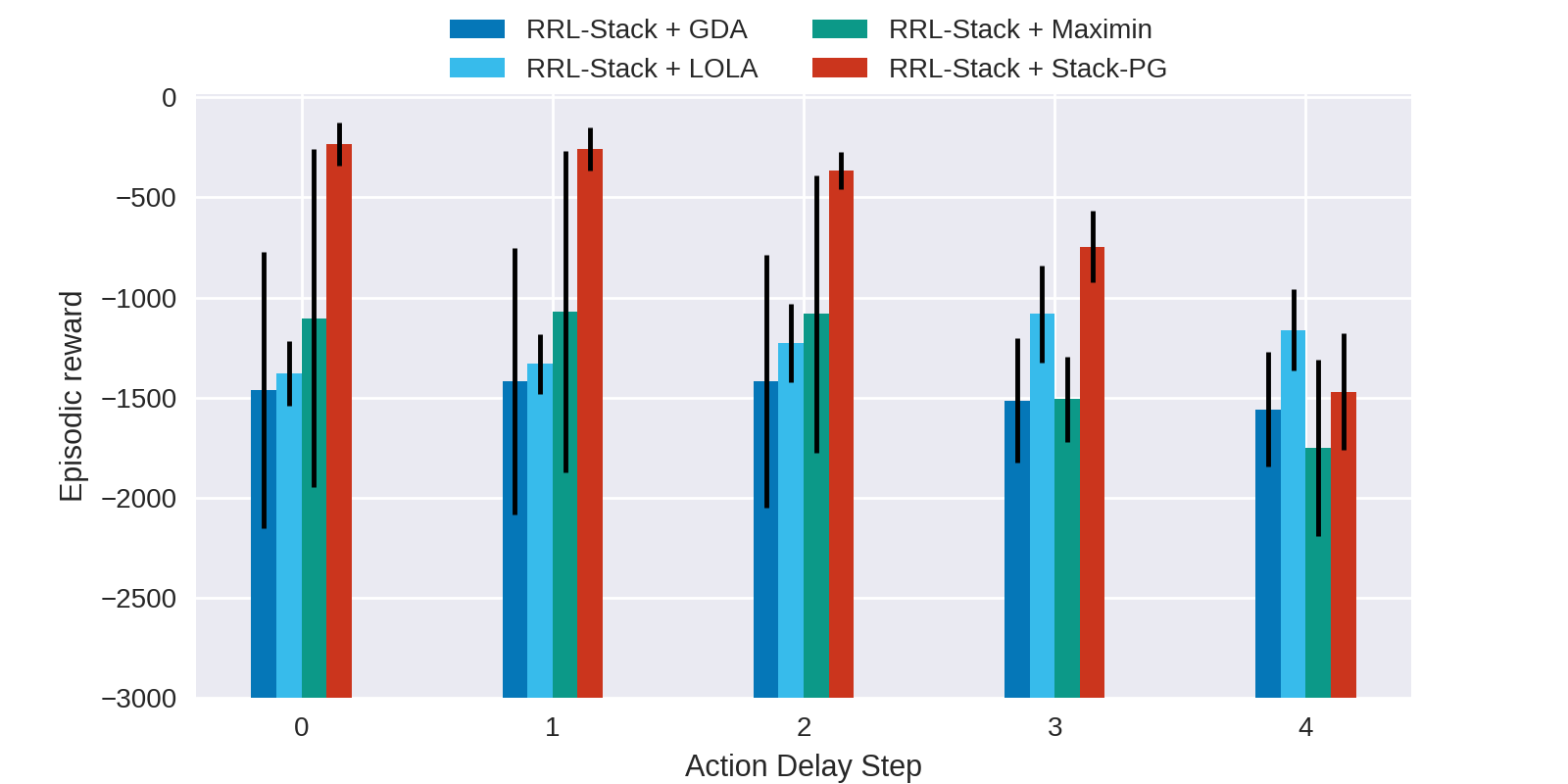}
  \vspace{-0.65cm}
  \caption{RRL-Stack formulation with different learning algorithms}\label{fig:LunarLanderRobustnessLD}
  \vspace{-0.50cm}
\end{figure}

To study whether Stack-PG helps stabilize the training, we use the same \textit{RRL-Stack} game formulation with $\alpha=0.5$ but apply different learning algorithms in Fig.~\ref{fig:LunarLanderRobustnessLD}. We test \textit{GDA}, \text{Maximin} and \textit{Stack-PG} and observe that \textit{Stack-PG} not only stabilizes the training process but also reduces the variance of performance significantly compared with \textit{GDA} and \text{Maximin}, which again answers \textbf{(Q2)}. It is consistent with recent works that have shown that opponent-aware modeling improves the training process stability in Generative Adversarial Networks \cite{schafer2019implicit}.

% \vspace{-0.10cm}

\section{Conclusion}
In this work, we study robust reinforcement learning via adversarial training problems. To the best of our knowledge, this is the first work to formalize the sequential nature of deployments of robust RL agents using the Stackelberg game-theoretical formulation. We enable the agent to learn robust policies in progressively challenging environments with the adaptively-regularized adversary. We develop a variant of policy gradient algorithms based on the Stackelberg learning dynamics. In our experiments, we evaluate the robustness of our algorithm on two tasks and demonstrate that our algorithm clearly outperforms the robust and non-robust baselines in single-agent and multi-agent tasks.

\section*{Ethical Statement}
There are no ethical issues.

\section*{Acknowledgments}
We gratefully acknowledge support from the National Science Foundation under grants IIS-1849304, CAREER CNS-2047454, and CAREER IIS-2046640.

\bibliographystyle{named}
\bibliography{ijcai22}

\end{document}